\documentclass[%
 reprint,
superscriptaddress,
preprintnumbers,
 amsmath,amssymb,
 aps,
]{revtex4-2}

\usepackage{bibunits}
\usepackage{setspace}

\usepackage{graphicx}
\usepackage{dcolumn}
\usepackage{bm}

\renewcommand{\v}[1]{\ensuremath{\mathbf{#1}}} 
 
\newcommand{\abs}[1]{\left| #1 \right|} 
 
\renewcommand\eqref[1]{Eq.\;\ref{#1}} 
\newcommand{\equalcontrib}{\thanks{These authors contributed equally to this work.}}

\usepackage [english]{babel}
\usepackage [english = american]{csquotes}
\MakeOuterQuote{"}

\usepackage{color}					

\begin{document}
\begin{bibunit}

\title{Transformers for dynamical systems learn transfer operators in-context}

\author{Anthony Bao}\equalcontrib
\affiliation{%
Department of Electrical \& Computer Engineering, The University of Texas at Austin, Austin, Texas 78712, USA
}
\author{Jeffrey Lai}\equalcontrib 
\affiliation{%
The Oden Institute, The University of Texas at Austin, Austin, Texas 78712, USA
}
\author{William Gilpin}
\email{wgilpin@utexas.edu}
\affiliation{%
Department of Physics, The University of Texas at Austin, Austin, Texas 78712, USA
}

\date{\today}

\begin{abstract}
Large-scale foundation models for scientific machine learning adapt to physical settings unseen during training, such as zero-shot transfer between turbulent scales.
This phenomenon, in-context learning, challenges conventional understanding of learning and adaptation in physical systems.
Here, we study in-context learning of dynamical systems in a minimal setting: we train a small two-layer, single-head transformer to forecast one dynamical system, and then evaluate its ability to forecast a \textit{different} dynamical system without retraining. 
We discover an early tradeoff in training between in-distribution and out-of-distribution performance, which manifests as a secondary double descent phenomenon.
We discover that attention models apply a transfer-operator forecasting strategy in-context. They (1) lift low-dimensional time series using delay embedding, to detect the system’s higher-dimensional dynamical manifold, and (2) identify and forecast long-lived invariant sets that characterize the global flow on this manifold.
Our results clarify the mechanism enabling large pretrained models to forecast unseen physical systems at test time without retraining, and they illustrate the unique ability of attention-based models to leverage global attractor information in service of short-term forecasts.
\end{abstract}

\maketitle

\clearpage

Scientific foundation models are pretrained on vast amounts of data, with a goal of discovering generalizable patterns applicable to unseen problem domains. 
Animating such efforts is the observation that, at sufficient scale, these models exhibit the emergent capability to solve unseen tasks during testing \cite{subramanian2023towards}. For example, models trained on low-resolution snapshots of turbulence exhibit the spontaneous ability to forecast flows on higher-resolution grids \cite{mccabe2024multiple,rahman2024pretraining,herde2024poseidon}. In some cases, models trained on certain classes of PDEs, like diffusion equations, develop the ability to forecast new PDEs, like wave equations \cite{mccabe2025walrus}. These findings align with similar observations for large language models, which exhibit the emergent ability to solve unseen tasks, like logical puzzles or mathematical derivations, at sufficient scale \cite{brown2020language}.

The underlying mechanism, in-context learning, differs from classical machine learning approaches that dichotomize learning during training from inference during testing. 
Instead, pretraining provides a sufficiently large model with vast background experience and, in some cases, a mechanistic world model, allowing it to adapt to new tasks based on minimal context \cite{von2023transformers,akyurek2023learning}. 
In traditional scientific machine learning, a trained model acts as a surrogate for the dynamical process that generated the training data, and so a trained model can, at best, only generate test outputs matching the distribution of the training data, known as \textit{in-distribution (ID) generalization}. 
In contrast, in-context learning enables \textit{out-of-distribution (OOD) generalization} on new data distributions \cite{goring2024out}. 
However, the underlying mechanisms enabling in-context learning and OOD generalization in SciML are poorly understood, even as recent empirical studies have uncovered robust signatures of generalization in areas as diverse as turbulence closure modelling, molecular dynamics, and magnetohydrodynamics \cite{clavier2025generative,schreiner2023implicit,rosofsky2023magnetohydrodynamics}.

To clarify the origin of in-context learning in scientific machine learning, we draw inspiration from recent works that train minimal language models, in order to probe the mechanistic origin of in-context learning \cite{garg2022can,reddy2024mechanistic,edelman2024evolution}. We train a small transformer model on a univariate trajectory originating from a single dynamical system (denoted Train-ID), and we then evaluate its ability to both (1) forecast a different trajectory from the \textit{same} dynamical system (Test-ID), and (2) forecast a trajectory from a \textit{different} dynamical system (Test-OOD). We first generate a trajectory $\v{y}(t) \in \mathbb{R}^D$ from the $D$-dimensional ordinary differential equation $\dot{\v{y}} = \v{f}(\v{y})$, $\v{y}(0)= \v{y}_0^\text{train}$. 
Train-ID consists of a univariate training time series $x_{1}, x_{2}, ..., x_T$ representing evenly-spaced samples of the first coordinate of the full trajectory $\v{y}(t)$. In order to use language modeling approaches, we convert these continuous values to discrete tokens by uniformly quantizing them into $V$ evenly-spaced bins (token IDs). This same approach is used by leading foundation models for zero-shot time series forecasting \cite{ansari2024chronos}. For simplicity, hereafter we use $x_t$ to refer to elements of the tokenized univariate time series.

A probabilistic forecast model $\hat{p}(x_{t+1} | \v{x}_C) \in \mathbb{R}^V$ accepts the context vector $\v{x}_{C} \equiv x_t, x_{t-1},\ldots,x_{t-C+1} \in \mathbb{Z}^C$ comprising $C$ past tokens, and returns a vector $\hat{p} \in \mathbb{R}^V$ corresponding to next-token probabilities across the vocabulary. We estimate $\hat{p}$ from data using a minimal GPT-like transformer, comprising two layers with a single attention head each, and relative positional encoding \cite{brown2020language,garg2022can,reddy2024mechanistic,edelman2024evolution} (\ref{app:architecture}). 
For each input context $\v{x}_C$, transformers compute self-attention, comprising $C^2$ pairwise comparisons among all tokens in the context, resulting in attention weights $A(\v{x}_C) \in \mathbb{R}^{C \times C}$ that quantify the relative importance of each past token in influencing the next-token probabilities.

We train the transformer $\hat{p}(x_{t+1} | \v{x}_C)$ on Train-ID using standard gradient-based optimization with cross-entropy loss, which aligns $\hat{p}(x_{t+1} | \v{x}_C) \in \mathbb{R}^V$ with the empirical next-token distribution $p(x_{t+1} | \v{x}_C)$ in the training data. 
After training, multistep forecasts are generated by sampling tokens from the transformer $\hat{x}_{t+1} \sim \hat{p}(x_{t+1} | \v{x}_C)$ and appending each sample to the context autoregressively.
Throughout training, we measure the transformer's ability to forecast an unseen trajectory from the same dynamical system (Test-ID) denoted $\dot{\v{y}} = \v{f}(\v{y})$, $\v{y}(0)=\v{y}_0^\text{test}$ as well as its ability to forecast a trajectory from an unseen system (Test-OOD) denoted $\dot{\v{y}} = \v{g}(\v{y})$, $\v{y}(0)=\v{y}_0$. We repeat all experiments by training new transformers on each of $100$ different pairs of systems drawn from a large database of ordinary differential equations \cite{gilpin2021chaos}, allowing measurement across a variety of distinct physical systems.

\begin{figure}
{
\centering
\includegraphics[width=\linewidth]{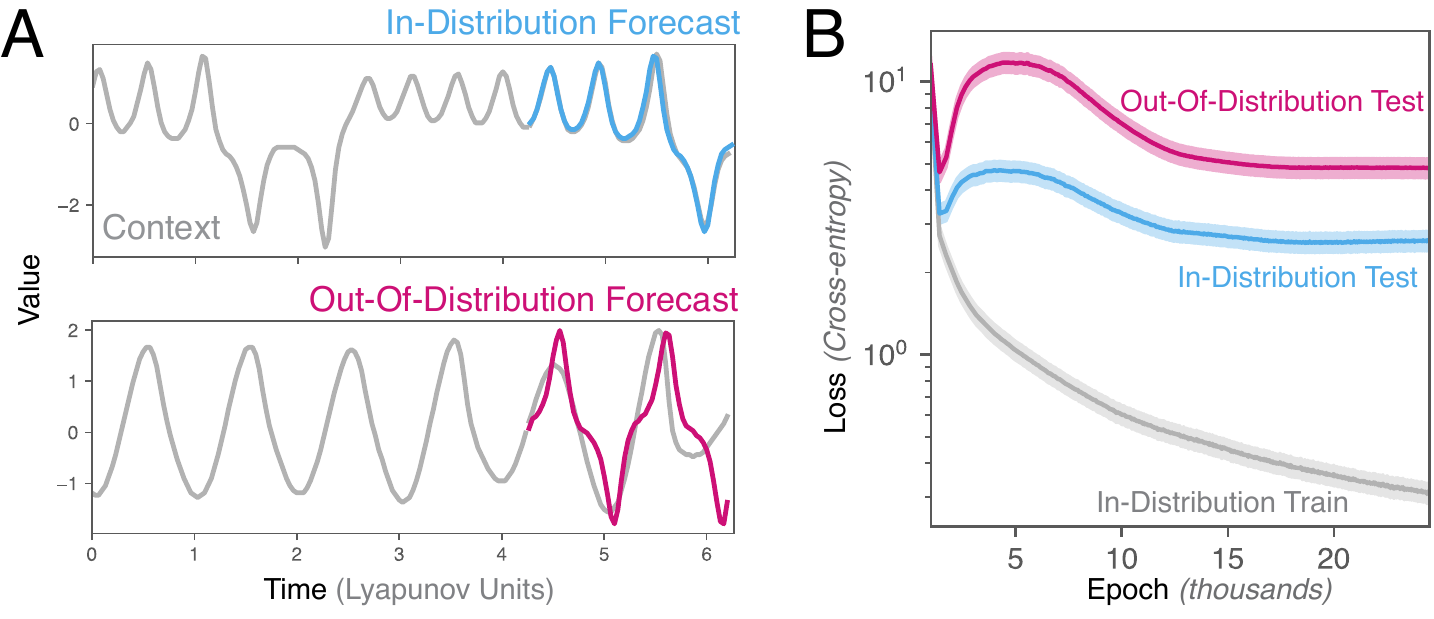}
\caption{
\textbf{Double descent during in-context learning of dynamical systems.} 
(A) Forecasts from a transformer trained on univariate time series from one dynamical system (Train-ID) then evaluated on its ability to forecast unseen trajectories from the same system (Test-ID, blue), versus forecasts of trajectories from an unseen system (Test-OOD, magenta). Gray curve shows the context, a subset of the total test data.
(B) Loss curves on the training data, versus held-out in-distribution and out-of-distribution validation sets.
Ranges are standard errors across $100$ replicate experiments, each corresponding to a different randomly-sampled train/test pair of dynamical systems.
}
\label{fig:learning}
}
\end{figure}

Remarkably, we observe that, in this minimal setting, small pretrained transformers exhibit the ability to predict unseen dynamical systems (Fig. \ref{fig:learning}A). These forecasts outperform naive baselines like mean regression, suggesting that transformers employ a more sophisticated in-context forecast strategy by leveraging their uniquely long context. 
During training on a single system (Train-ID), we see an early underfitting-to-overfitting transition on Test-ID, consistent with the classical bias-variance tradeoff: the model initially learns generalizable information about the system's dynamics, leading to improvements on both seen and unseen data (Fig. \ref{fig:learning}B, gray and blue curves). However, as the Train-ID loss continues to fall, the model begins to overfit, leading to an increase in Test-ID error. At even later stages of training, we observe epoch-wise double descent, a recently-characterized phenomenon in which the Test-ID error further falls again below the classical bias-variance optimum \cite{belkin2019reconciling,rocks2022memorizing,nakkiran2021deep}. However, we discover a \textit{second}, earlier double descent curve in the Test-OOD error (Fig. \ref{fig:learning}B, magenta curve). This finding implies that information generalizable to other dynamical systems becomes available to the transformer during the pretraining process, but that it is bounded from above by ID-system-specific information.

We quantify OOD generalization by measuring how Test-OOD error scales with the statistical distance between Train-ID and Test-OOD. Across hundreds of train-test system pairs, we compute the transformer OOD error as the cross-entropy of next-token predictions averaged across Test-OOD $\mathcal{L}_{\mathrm{OOD}} = \mathcal{H}(p(x_{t+1} | \v{x}_C),\; \hat{p}(x_{t+1} | \v{x}_C))$. The cross-entropy may be decomposed into two terms $\mathcal{H}(p, \hat{p}) = \mathcal{H}(p) + D_\text{KL}(p || \hat{p})$. Because the model is trained on Train-ID, we expect that the second term is proportional to the distribution shift between Train-ID and Test-OOD. We estimate this term by separately fitting $k$-order Markov models on Train-ID and Test-OOD $p^\text{ID}_\text{k-markov}(x_{t+1} | \v{x}_k)$ and $p^\text{OOD}_\text{k-markov}(x_{t+1} | \v{x}_k)$, $\v{x}_k \equiv x_{t},\ldots,x_{t-k+1}$. We compute the KL divergence between these Markov surrogates to obtain the approximate relationship
\begin{equation}
\mathcal{L}_\text{OOD}
\approx
\mathcal{H}\!\left(p^\text{OOD}_{k\text{-markov}}\right)
+
D_{\mathrm{KL}}\!\left(
p^{\text{OOD}}_{k\text{-markov}}
\,\middle\|\,
p^{\text{ID}}_{k\text{-markov}}
\right)
\label{eq:lood}
\end{equation}
This equation thus relates the error of the transformer on a new dynamical system to the degree of train/test shift.

\begin{figure}
{
\centering
\includegraphics[width=\linewidth]{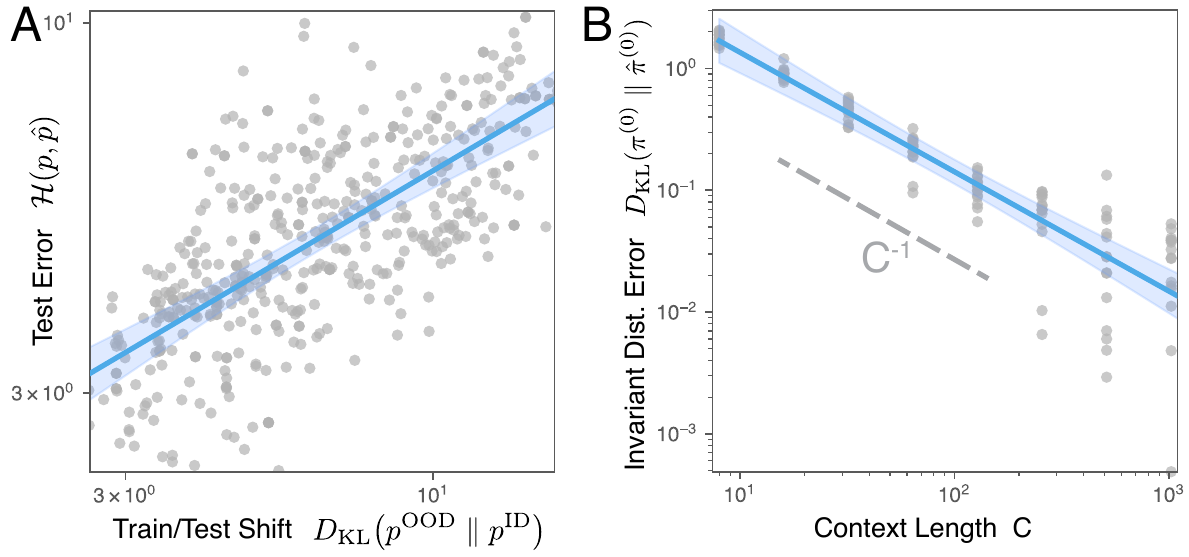}
\caption{
\textbf{Scaling laws for out-of-distribution generalization in dynamical systems.}
(A) The test error of the unseen system (cross-entropy, Test-OOD) versus the difference between the training and testing sets (KL divergence between the attractor of Test-OOD and Train-ID).
(B) The error in the steady-state invariant distribution of the transformer dynamics relative to the true invariant distribution of Test-OOD, versus the context length $C$. Dashed line shows the expected scaling for an estimator that counts observed state transitions in-context \cite{edelman2024evolution}.
Points indicate $500$ transformers trained and tested on distinct ID/OOD dynamical system pairs, and solid lines indicate linear fits with 99\% confidence intervals.
}
\label{fig:scaling}
}
\end{figure}

When sufficient data is available the first term in \eqref{eq:lood} plateaus in $k$. However, finite context and training data limit the accuracy of the Markov description at large $k$ and so, for our experimental setting, we find that the bigram ($k=2$) case exhibits the strongest relationship. Across our experiments, we find that the OOD loss is well-described by the expected linear scaling $\mathcal{L}_{\mathrm{OOD}} \propto D_{\mathrm{KL}}^{(2)}$ in \eqref{eq:lood} (Fig. \ref{fig:scaling}A, Spearman's $\rho=0.72 \pm 0.02$, $500$ system pairs). 
Our observations align with domain-adaptation theory, which predicts that target risk depends on the task-relevant difference between the source and target distributions \cite{ben2010theory,mansour2009domain}.

\begin{figure}
{
\centering
\includegraphics[width=\linewidth]{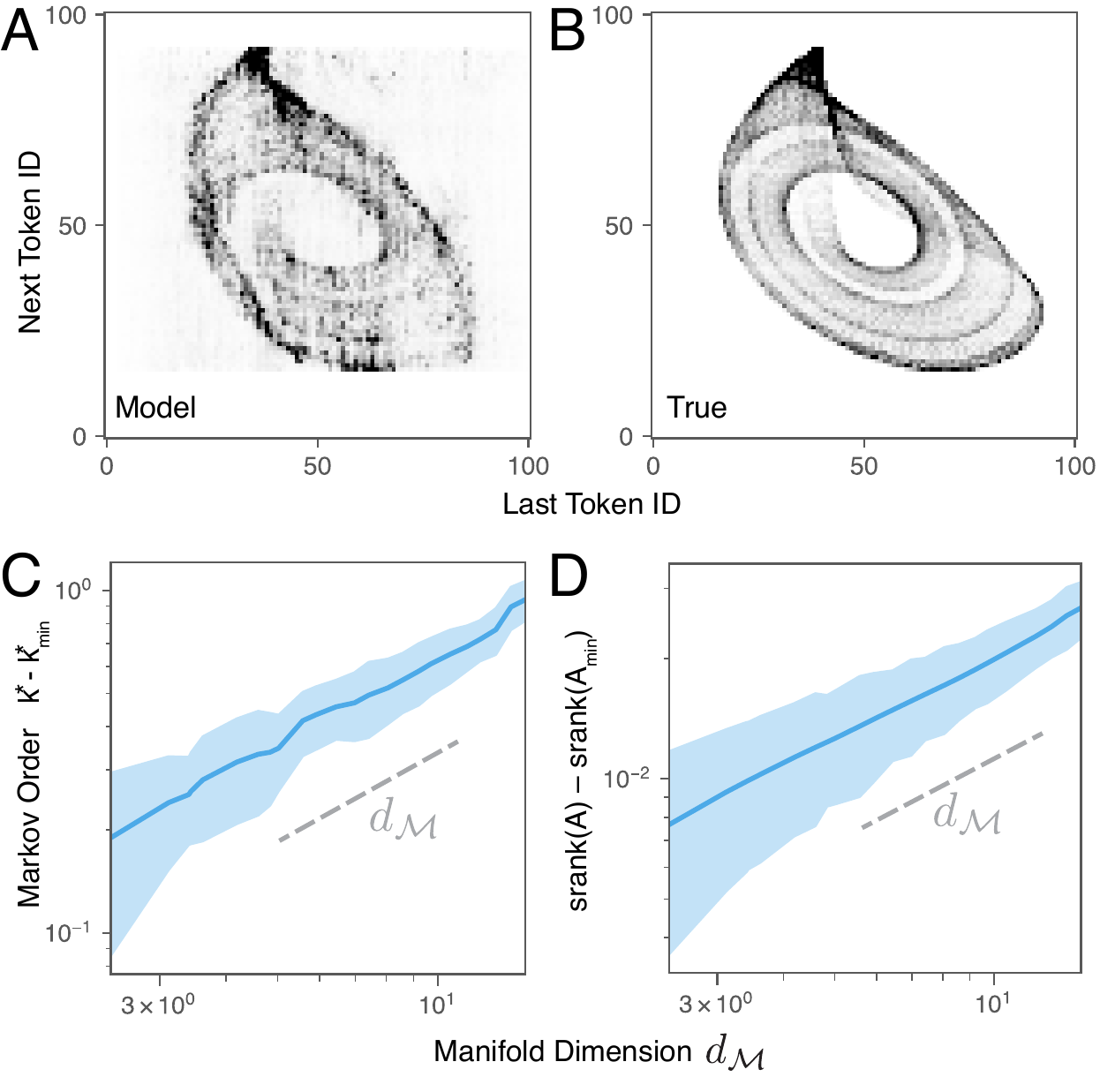}
\caption{
\textbf{Transformers perform time-delay embedding during inference.}
(A) Empirical time-delayed next-token probabilities $\hat{p}(x_{t+1} | x_{t-k})$ averaged across Test-OOD context for a transformer trained on a different system (Train-ID). 
(B) Exact next-token probabilities $p_\text{true}(x_{t+1} | x_{t-k})$ obtained from fitting a Markov chain on a long sample of Test-OOD.
(C) The order of the closest-approximating Markov chain, versus the true intrinsic dimension $d_\mathcal{M}$ of the full attractor from which the univariate Test-OOD trajectory originates.
(D) The effective dimension (stable rank) of the transformer's attention rollout matrix versus the true intrinsic dimension of Test-OOD. Stable rank is scaled by the maximum possible rank.
Annotated slopes indicate the linear scaling predicted by Takens' theorem.
Each point is an average over $100$ replicate models trained on a different low-dimensional dynamical system.
}
\label{fig:embedding}
}
\end{figure}

What strategies allow the transformer to generalize?
We sample the transformer's time-delayed conditional distribution $\hat{p}(x_{t+1} | x_{t-k})$, $k\leq C$ averaged across Test-OOD for all lag values $k \in 0, 1, \ldots, C$. We find that this distribution qualitatively resembles the true attractor of Test-OOD, implying that the univariate model identifies relevant information about the unobserved multivariate attractor during testing (Fig. \ref{fig:embedding}A).
This result is consistent with Takens' theorem: if a univariate time series arises from an attractor with intrinsic dimension $d_\mathcal{M}$, embedding the time series with at least $d_E = 2 d_\mathcal{M}+1$ time delays recovers the key geometric features of the true attractor \cite{takens1981dynamical}. We denote the time delay coordinates $\hat{\v{y}}_i = [x_{i-d_E+1}, ..., x_i]^\top$ as estimates of the unobserved true phase space coordinates $\v{y}$ \cite{botvinick2025invariant}.

To test the scaling predicted by Takens' theorem, we generate new Test-OOD trajectories using the Lorenz-96 system, a dynamical system with a controllable number of dynamical variables $D$ \cite{lorenz1996predictability}. This system's manifold dimension $d_\mathcal{M}$ scales approximately linearly with $D$ (\ref{app:gpdim}).
As we vary the number of dynamical variables in Test-OOD $D=3,4,\ldots,25$, we evaluate the average Test-OOD accuracy of transformers originally trained on a different dynamical system (Train-ID) with $D=3$.
For each test case, we compare the transformer's empirical $k$-order conditional distribution (averaged across Test-OOD) to a $k$-order Markov chain $p_\text{k-markov}^\text{OOD}(x_{t+1} | \v{x}_k)$ trained directly on Test-OOD, in order to estimate the effective order $k$ of the Markov chain best-approximating the transformer's Test-OOD dynamics. 
We find that the order of the best-approximating Markov chain scales linearly with the intrinsic dimension of Test-OOD (Fig. \ref{fig:embedding}C), consistent with Takens' theorem. This finding is striking, because the univariate Test-OOD series contains no direct information about the true dimensionality $D$ or $d_\mathcal{M}$, suggesting that the transformer instead adapts during inference by estimating the intrinsic dimensionality of the input time series.

To confirm that this effect arises from the transformer's internal representations, we calculate the attention rollout matrix $A_\text{roll} \in \mathbb{R}^{C \times C}$ \cite{abnar2020quantifying}. This metric measures the context-dependent response of the model to different token positions, averaged across the entire testing set and both attention heads.
We compute the stable rank of the rollout matrix ($d_\text{latent} = \text{srank}(A_\text{roll})$) as a measure of the intrinsic dimensionality of the transformer's latent representations.
We find that increasing attractor dimension $d_\mathcal{M}$ linearly correlates with increasing latent dimensionality $d_\text{latent}$. 
Because the univariate time series presented to the transformer lack direct information about dimensionality, we conclude that the trained model infers information about the underlying attractor via in-context learning.
Thus, not only does the transformer embed the test time series, it does so adaptively by identifying when a univariate time series arises from a high-dimensional system.

\begin{figure*}
{
\centering
\includegraphics[width=\linewidth]{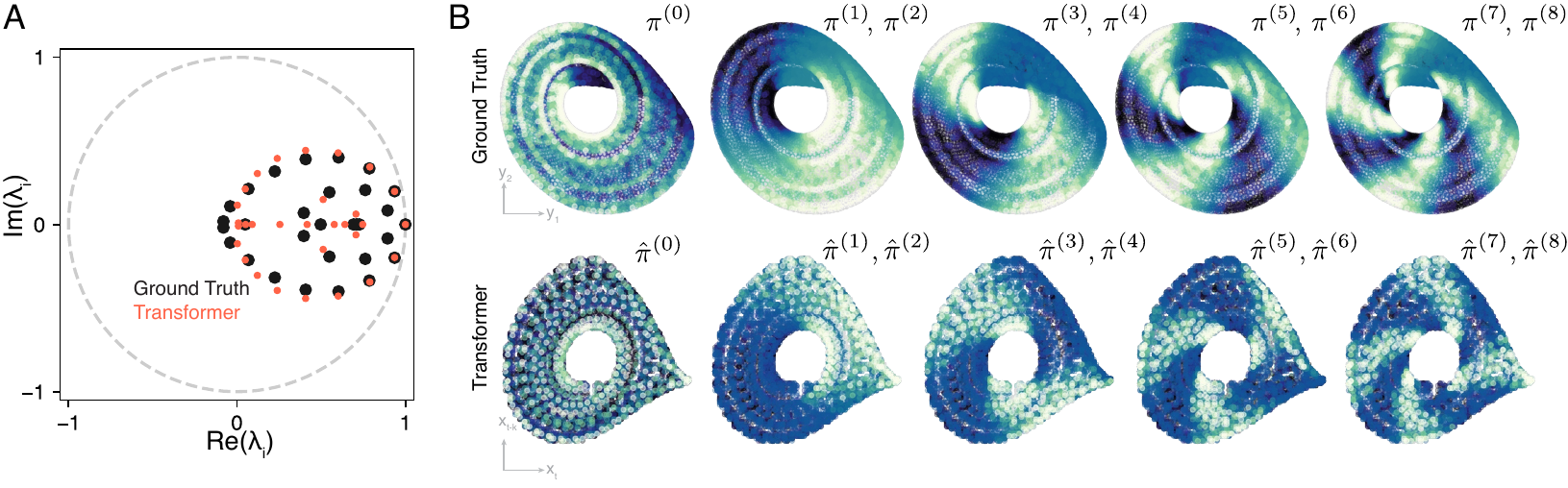}
\caption{
\textbf{Transformers learn in-context transfer operators on reconstructed dynamical manifolds.}
(A) The eigenvalue spectrum of the transfer operator estimated from the fully-observed Test-OOD attractor $p(\v{y}_{t+1} | \v{y}_t)$, and the time-lagged transfer operator estimated by sampling contiguous length-$k$ sequences from the transformer $\hat{p}(\hat{\v{y}}_{t+1} | \hat{\v{y}}_{t}) = \hat{p}(x_{t+1}, x_t, ... x_{t-k} | x_t, x_{t-1}, ... x_{t-k-1})$.
(B) (Top) The invariant distribution $\pi^{(0)}(\v{y})$ and longest-lived metastable distributions ($\pi^{(i)}(\v{y})$, $i>0$) of the true transfer operator, ranked by $\abs{\lambda_i}$.
(Bottom) The invariant and metastable distributions of the transformer, based on its implied transfer operator $\hat{p}(\hat{\v{y}}_t | \hat{\v{y}}_{t-1})$.
Because the transformer is univariate, eigenvectors are plotted on the time-delay embedding $\hat{\v{y}} \equiv [x_{t}, x_{t-1}]^\top$.
}
\label{fig:operator}
}
\end{figure*}

We next consider how the model propagates the dynamics after estimating the dynamical coordinates $\hat{\v{y}}$.
Broadly, there are two ways that a surrogate model could represent a dynamical propagator, given finite-length context: (1) a differential form based on approximating time derivatives (as in multi-step integrators, or differential equation inference methods like SINDy), or (2) an integral approach that approximates a transfer operator (as in neural operators or discrete-time system identification) \cite{buzhardt2025relationship,brunton2022data}.
Based on the coincidence of the trained model's empirical next-token distribution with a Markov chain, we hypothesize the latter: transformers approximate propagators in-context.

To test this possibility, we extract the empirical transition matrix of the transformer on the space of length-$k$ delay vectors $\hat{\v{y}}_t$ by aggregating multi-step predictions from the model $\hat{p}(x_{t+1}\mid \v{x}_C)$. Briefly, across Test-OOD we tabulate each distinct $k$-gram ($\v{x}_k$, $k<C$), and we group together all context sequences $\v{x}_C$ that end in that $k$-gram. For each unique $k$-gram, we randomly-sample from all matching context sequences and predict the frequency of the next token $x_{t+1}$. By repeating this process separately for each $k$-gram, we estimate a transfer operator on the time-delayed sequence $\hat{p}(\v{\hat{y}}_{t+1}\mid \v{\hat{y}}_t) \equiv \hat{p}(x_{t+1},x_t,...,x_{t-k} | x_t,x_{t-1},...,x_{t-k-1})$. 
This row-stochastic transition matrix describes the model-implied dynamics on the lag-$k$ state space embedding of the univariate Test-OOD time series. In dynamical systems theory, this transition matrix represents a coarse-grained estimate of the Perron-Frobenius operator, which describes how dynamical systems evolve distributions of trajectories over time \cite{cvitanovic2016chaos}.

We compare the empirical operator to the ground truth operator $p(\v{y}_{t+1}\mid \v{y}_t)$, which we estimate from the \textit{fully-observed} state space using Ulam's method: we discretize the full multivariate phase space and tabulate all transition rates between pairs of regions \cite{ulam1960collection}.
We observe that the leading eigenvalues (ranked by $\abs{\lambda_i}$) closely match, implying that the transformer captures the dominant timescales associated with the Perron-Frobenius operator propagating the dynamics (Fig. \ref{fig:operator}A). 
The corresponding eigenfunctions represent the stationary distribution $\pi^{(0)}$ and subleading eigenfunctions $\pi^{(1)}$, $\pi^{(2)}$ corresponding to long-lived metastable trapping regions, known as \textit{almost-invariant sets} \cite{froyland2003detecting}. These eigenfunctions occur in pairs due to symmetry (Fig. \ref{fig:operator}B).

Our findings agree with recent studies showing that minimal, two-layer transformers as we consider here can implement $k$-gram counting circuits in-context \cite{edelman2024evolution}, potentially enabling estimation of the transfer operator via Ulam's method. Given an estimate $\hat{p}(\v{y}_{t+1} | \v{y}_{t})$ of the true transfer operator ${p}(\v{y}_{t+1} | \v{y}_{t})$, the KL divergence scales quadratically to leading order $D_\text{KL}(p || \hat{p}) \sim \mathcal{O}((p -\hat{p})^2)$ when the discrepancy is small. Because the number of expected counts of a given state $i$ increases linearly in the context $N_i \approx \pi_i^{(0)} C$, bounds from eigenvalue perturbation theory predict $D_\text{KL}(p || \hat{p}) \sim \mathcal{O}(C^{-1})$ (\ref{app:klcontext}). We find that this scaling holds in practice across many models trained on distinct Train-ID/Test-OOD system pairs (Fig. \ref{fig:scaling}B). We thus conclude that, as the transformer's out-of-distribution performance improves, it learns in-context an increasingly accurate estimate of the Perron-Frobenius operator of the fully-observed true system.

Taken together, our results show that, during inference, transformers trained on dynamical systems first learn to perform time delay embedding on the context, and then use co-occurrence statistics to approximate the true transfer operator of the fully-observed state space. 
Our work thus provides an example of how a learning model can extract a system-independent forecasting strategy despite system-specific training, beyond naive strategies observed in prior works such as parroting sequences from the context \cite{zhang2026context}. 
Our work thus provides a potential explanation for recent works showing the surprising ability of pretrained foundation models to forecast dynamical systems outside their training data, such as turbulent flows, weather fronts, or plasma dynamics \cite{mccabe2024multiple,price2025probabilistic,rosofsky2023magnetohydrodynamics}.
Additionally, our observation of in-context time delay embedding supports findings throughout the broader time series literature, that univariate models often match the performance of multivariate models, even for datasets with strong covariation \cite{nie2023a}. 
It also accords with a recent theoretical result showing that single-layer transformers can implement adaptive time-delay embedding in-context, given sufficient latent dimensionality \cite{duthe2025mechanistic}.
Finally, our results support recent neural scaling laws showing improved performance of time series and scientific foundation models with additional context \cite{lai2026panda}.
Reminiscent of similar findings in language models \cite{zhang2024trained}, our work demonstrates how the unique structure of attention-based models enables nontrivial strategies for estimating and propagating dynamical states.

\putbib[cites]
\end{bibunit}

\clearpage
\newpage
\onecolumngrid
\section*{Appendix}
\addcontentsline{toc}{section}{Appendix}
\setcounter{page}{1} 
\renewcommand{\thetable}{S\arabic{table}}
\setcounter{table}{0}
\renewcommand{\thefigure}{S\arabic{figure}} 
\setcounter{figure}{0}
\renewcommand{\theequation}{A\arabic{equation}}
\setcounter{equation}{0}
\renewcommand{\thesubsection}{\Alph{subsection}}
\setcounter{subsection}{0}
\renewcommand{\thesection}{Appendix \Alph{section}}
\setcounter{section}{0}

\tableofcontents
\begin{bibunit}

\section{Code Availability}

The code used in this study is available at \url{https://github.com/williamgilpin/icicl}

\section{Model architecture and training}
\label{app:architecture}

\noindent\textbf{Architecture.}We implement a compact causal language model operating on a discrete vocabulary ($V=100$) that predicts the next token autoregressively. Our model is a decoder-only transformer, similar to GPT models but with relative positional embedding \cite{brown2020language,raffel2020exploring}. Each input token is mapped to a higher-dimensional embedding ($d_{\text{model}}=256$). The network consists of two identical transformer-style residual blocks, each using single-head causal self-attention, with query and key projections into $d_k=128$ dimensions, and a value projection back into $d_{\text{model}}$ \cite{vaswani2017attention}. Attention logits are scaled by $d_k^{-1/2}$ and strictly masked with a lower-triangular causal mask. Each block applies pre-normalization with LayerNorm before attention \cite{ba2016layer}, then adds the attention output residually, and then applies a second LayerNorm followed by a position-wise feed-forward network with expansion factor $2$ (resulting in hidden width $2d_{\text{model}}=512$), using ReLU activation and a second linear layer returning to $d_{\text{model}}$, again added residually. 
A final LayerNorm precedes the output projection to logits over the vocabulary. The output projection is weight-tied to the input embedding matrix. The maximum context length is a fixed block size of $C=512$ tokens. Positional information is encoded using ALiBi, an additive attention-logit bias equal to the nonnegative lag distance: $-(i-j)$ for $i\ge j$ and $0$ otherwise. 

\vspace{1em}\noindent\textbf{Tokenization.} We convert real-valued time series into tokens with mean scaling followed by uniform scalar quantization into $V$ bins, following the approach used in the first generation of the Chronos family of time series foundation models \cite{ansari2024chronos}.
We favor direct quantization over patch-based approaches used in other time series foundation models due to the interpretability of the token space, and because it allows us to design our study based on prior studies of in-context generalization in language models \cite{garg2022can,bao2026universal}.

Given training time series $x_{1:T}$ of length $T$, we compute a scale factor from the context
\[
    s=\max\!\left(\frac{1}{T}\sum_{i=1}^{T}\lvert x_i\rvert,\ \varepsilon\right),\qquad \varepsilon=10^{-8},
\]
and we then normalize the entire series $\tilde{x}_i=x_i/s$. The normalized values are discretized into integer bin IDs in $\{1,\dots,V\}$ using uniformly spaced bin centers $c_j$ spanning $[c_{\min},c_{\max}]$ (with $c_{\min}$ and $c_{\max}$ treated as hyperparameters), and decision boundaries placed at midpoints between adjacent centers, with outer edges extending to $\pm\infty$ so that all values map to a unique bin. A reserved "padding" token uses ID $0$. 
Missing values (NaNs) are handled by replacing their contribution with $0$ for purposes of scale computation, while their token positions are set to the padding ID $0$ in the emitted token sequence. 

\vspace{1em}\noindent\textbf{Training.} We train our model using next-token prediction with cross-entropy loss over the vocabulary. We quantize the training time series and provide it to the model as a single long one-dimensional token stream, from which we sample fixed-length windows of length $T=512$ uniformly at random: the input is a contiguous block of $512$ tokens and the target is the same block shifted by one position. The effective dataset length is treated as a large virtual size ($10,000$ samples) with random window sampling each time. We optimize the model using AdamW with learning rate $5\times 10^{-5}$, weight decay $0.0$, batch size $32$, with gradient accumulation over micro-batches (with default size $1$) \cite{loshchilov2017decoupled}. 
We clip gradients to a maximum global norm of $1.0$ at each step. We train for $50,000$ optimization steps and perform validation every $1000$ steps by evaluating cross-entropy validation batches. We do not use a learning-rate schedule.

\vspace{1em}\noindent\textbf{Inference.} We perform autoregressive inference for a specified number of new tokens $H$ beyond the provided context. At each step, we use only the most recent $C=512$ context tokens. The model outputs logits for the next token, which are converted to a token choice via sampling. \
We append the sampled token to the context, and repeat the process for $H$ steps, yielding a lengthened token sequence. 

\section{Estimating the fully-observed transfer operator with Ulam's method}
\label{sec:ulam}

Let $\{\v{y}_t\}_{t=0}^{N-1}$ be a time series (discrete-time) sampled at uniform intervals $\tau$ from a trajectory $\v{y}(t) \in \mathbb{R}^D$ in the fully-observed phase space. We construct a finite-rank approximation of the Perron-Frobenius (transfer) operator at lag $\tau>0$ by (1) partitioning the $D$-dimensional phase space into $K$ bins, (2) symbolizing the trajectory by bin labels, and (3) estimating the induced Markov transition matrix between bins.

\vspace{1em}\noindent\textbf{Partitioning.} We partition $\{B_1,\dots,B_K\}$ the data $\{\v{y}_t\}$ by clustering. Given a set of samples $\{\v{y}_t\}$ we use $K$-means clustering to generate Voronoi cells $B_i \equiv \{\v{y}:\arg\min_j\|\v{y}-\v{c}_j\| = i\}$ with centers $\{\v{c}_i\}_{i=1}^K$, $\v{c}_i \in \mathbb{R}^D$. This defines a coding map $\kappa:\v{y}\to\{1,\dots,K\}$, which we apply to the training data to generate the symbolic dynamics
\[
    s_t \;:=\; \kappa(\v{y}_t)\in\{1,\dots,K\}.
\]

\vspace{1em}\noindent\textbf{Estimating the Ulam matrix.} For a fixed lag $\tau$, Ulam's method approximates the transfer operator $\mathcal{P}_\tau$ by the $K\times K$ matrix $P^{(\tau)}$ as
\[
    P^{(\tau)}_{ij} \;\approx\; \mathbb{P}(s_{t+\tau}=j \mid s_t=i),
\]
i.e., the probability mass transferred from bin $B_i$ to bin $B_j$ over time $\tau$. We compute this quantity by counting all transition pairs separated by $\tau$, and dividing by the total number of transitions observed exiting the symbol associated with each row.

\vspace{1em}\noindent\textbf{Estimating the transfer operator spectrum.} Once $P^{(\tau)} \in \mathbb{R}^{K \times K}$ is obtained, we treat it as a Markov operator. We estimate the stationary distribution $\pi^{(0)}$ from the invariant left eigenvector of $P^{(\tau)}$,
\[
    \pi^\top P^{(\tau)} = \pi^\top,\qquad \sum_{i=1}^K \pi_i = 1,\qquad \pi_i\ge 0.
\]
We solve this using power iteration on $(P^{(\tau)})^\top$.

We estimate long-lived transient modes $\pi^{(i)}$, $i>0$ by applying power iteration to a deflated operator that removes the stationary mode. Writing $\mathbf{1}$ for the all-ones vector, the deflation
\[
    \widetilde{P}^{(\tau)} \;=\; P^{(\tau)} - \mathbf{1}\,\pi^\top
\]
annihilates the stationary left eigenvector and allows us to estimate the leading subdominant left modes (metastable transients) with power iteration.

\vspace{1em}\noindent\textbf{Choosing the partition size.} Because Ulam discretization trades bias (too few bins) against variance/sparsity (too many bins), we scan over candidate $K$ and select a value that yields informative, non-degenerate transition structure. We compute a row-wise entropy statistic of $P^{(\tau)}$ across $K$ and choose values that maximize this statistic. This heuristic favors partitions that are neither so large that most transitions are self-loops, nor so small that the transition matrix becomes overly sparse.

\section{Variation of model properties with context length}

\begin{figure*}
{
\centering
\includegraphics[width=\linewidth]{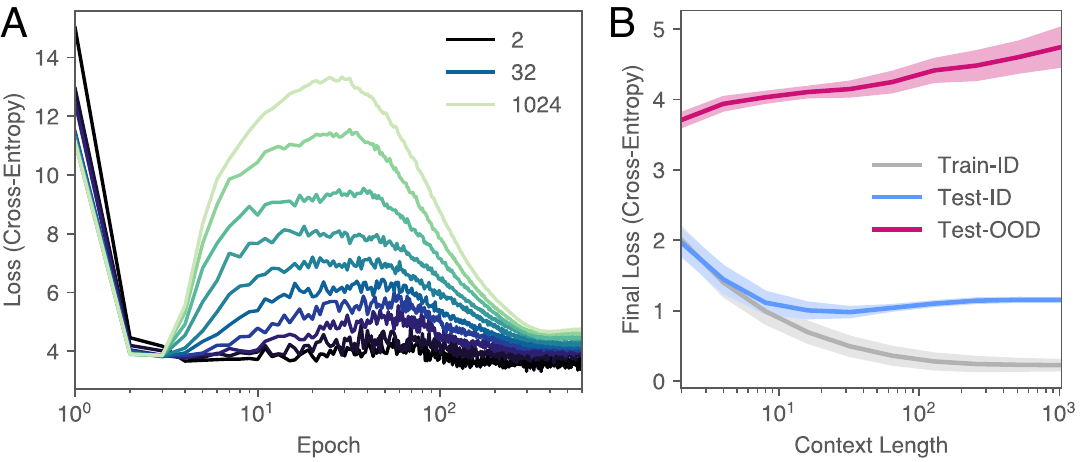}
\caption{
(A) Test-OOD loss curves for replicate models trained with varying context lengths $C$.
(B) The final cross-entropy loss for training data, in-distribution test data, and out-of-distribution test data, as a function of context length $C$.
}
\label{fig:context}
}
\end{figure*}

To probe the dependence of the model's properties on context length, we retrain the model with varying context lengths $C$. We find that the epochwise double-descent phenomenon that we report in the main text becomes more pronounced at longer context lengths (Fig. \ref{fig:context}), consistent with existing results for increasing model complexity on a finite training dataset \cite{nakkiran2021deep}. Comparing the final training and validation losses across models, the training and in-distribution test losses both decrease with increasing context, though the latter plateaus. The out-of-distribution test loss plateaus early, implying that the simplistic transfer operator strategy employed by the model in-context does not benefit from additional context during pretraining. We expect that this occurs because only low-order $k$-gram statistics are needed to estimate the transfer operator.

\section{Estimating the manifold dimension of the Lorenz-96 system}
\label{app:gpdim}

We consider the $D$-dimensional Lorenz-96 system with constant forcing,
\begin{equation}
    \dot{X_i}
    = (X_{i+1} - X_{i-2}) X_{i-1} - X_i + F,
    \qquad i = 1, \dots, D,
\end{equation}
with cyclic boundary conditions $X_{i+D} = X_i$ and $X_{i-D} = X_i$. Here $X_i(t)$ are the state variables, $D$ is the system dimension, and $F$ is the constant external forcing. In all simulations we set $F=8.0$. The initial condition is taken as $X_i(0)=F$ for all $i$, with a random normal perturbation of $10^{-2}$ added to the first component to ensure diversity across generated trajectories. The system is integrated to $1000$ time units using the implicit Radau IIA method. We discard the first 500 steps of the generated trajectory as transients.

Given a multivariate time series $\{ \v{y}_i \}_{i=1}^{N}\subset\mathbb{R}^D$, the Grassberger-Procaccia calculation is implemented by first forming the full matrix of pairwise Euclidean distances $\rho_{ij}=\lVert \v{y}_i-\v{y}_j\rVert_2$ and collecting all strictly positive distances into a single sample $\{\rho_k\}_{k=1}^{K}$ \cite{grassberger1983measuring}. A power law is fit to this empirical distribution using by the continuous maximum-likelihood formula for fitting power laws \cite{clauset2009power}. This scaling exponent is interpreted as the manifold dimension $d_\mathcal{M}$.

\begin{figure*}
{
\centering
\includegraphics[width=0.8\linewidth]{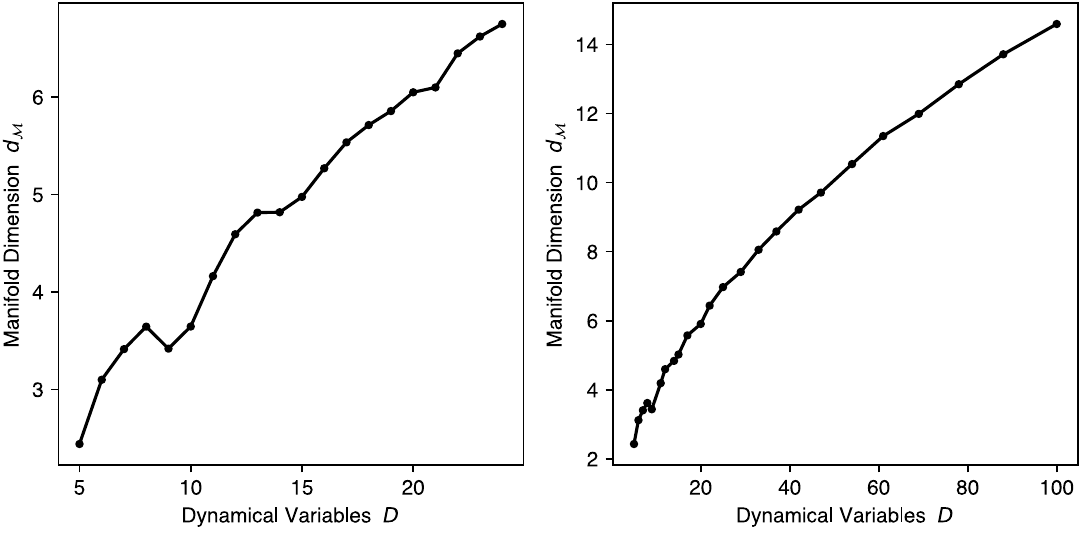}
\caption{
The estimated manifold dimension $d_\mathcal{M}$ of the attractor of the Lorenz-96 system as the number of dynamical variables $D$ varies.
}
\label{fig:gpdim}
}
\end{figure*}

To measure the scaling of the manifold dimension $d_\mathcal{M}$ with the dynamical dimension $D$, we generate trajectories from the Lorenz-96 system with varying number of dynamical variables $D = 5,6,7...,25$, and perform the Grassberger-Procaccia algorithm separately for each trajectory (Fig. \ref{fig:gpdim}). We observe a linear scaling of the manifold dimension $d_\mathcal{M}$ with the dynamical dimension $D$.

\section{Attention rollout}
\label{app:rollout}

For a fixed context length $C$ and an input token sequence $\v{x}_C\in\mathbb{Z}^C$, let $A^{(\ell)}(\v{x}_C)\in\mathbb{R}^{C\times C}$ denote the causal self-attention matrix produced by layer $\ell\in\{1,\dots,L\}$ (though we set $L=2$ in our experiments), where each row is a probability distribution over past positions (row-stochastic under the softmax and causal mask). Following \cite{abnar2020quantifying}, we incorporate residual mixing by defining the augmented layer-wise transition
\begin{equation}
    \widetilde{A}^{(\ell)}(\v{x}_C)\;=\;I + A^{(\ell)}(\v{x}_C),
\end{equation}
and propagate attention through depth by matrix multiplication,
\begin{equation}
    A_{\mathrm{roll}}(\v{x}_C)\;=\;\widetilde{A}^{(L)}(\v{x}_C)\,\widetilde{A}^{(L-1)}(\v{x}_C)\cdots \widetilde{A}^{(1)}(\v{x}_C)\;\in\;\mathbb{R}^{C\times C}.
\end{equation}
The entry $(A_{\mathrm{roll}})_{ij}$ can be interpreted as the effective contribution (via all compositional attention paths across layers) of token position $j$ to the representation at position $i$. In practice, we average $A_{\mathrm{roll}}(\v{x}_C)$ across the Test-OOD contexts to obtain a single rollout matrix $A_{\mathrm{roll}}\in\mathbb{R}^{C\times C}$ summarizing how the trained model distributes influence across lags; we then quantify its effective dimensionality using the stable rank $d_{\mathrm{latent}}=\mathrm{srank}(A_{\mathrm{roll}})=\|A_{\mathrm{roll}}\|_F^2/\|A_{\mathrm{roll}}\|_2^2$.

\section{In-context learning of transfer operators}
\label{app:klcontext}
Our results follow from a recent study that shows the ability of two-layer transformers to estimate Markov chains in-context \cite{edelman2024evolution}: in a dynamical systems setting, transfer operator estimation corresponds to bigram counting on delay coordinates.
For a Markov-chain prediction task with a Dirichlet prior over transition rows, the Bayes-optimal in-context predictor is the Dirichlet-multinomial posterior mean. Prior work shows that a 2-layer single-head, attention-only transformer can implement the bigram-counting circuit needed to estimate this mean \cite{edelman2024evolution}. 

Let $x_t \in \{1, 2, ..., \}$ be the quantized scalar time series. For an embedding length $m$, we define the delay state $s_t = (x_t,x_{t-1},\dots,x_{t-m+1}) \in \mathcal{S}$. We assume $m$ is large enough that the delay-state process is an approximately first-order Markov chain
\[
    \Pr(s_{t+1}\mid s_t,s_{t-1},\dots) \approx \Pr(s_{t+1}\mid s_t) = P_{s_t,s_{t+1}},
\]
where $P$ is a row-stochastic transition matrix on $\mathcal{S}$, which consists of $M$ distinct states. In order to model the dynamics, the transformer seeks to construct an approximation of this operator $\hat{P}$ given finite-length context.

For a given context, a one-step forecast reduces to estimating the row $p_{i}$ associated with the current delay state $s_t=i$.
\[
    \Pr(s_{t+1}\mid s_t=i)=p_i.
\]
The model seeks to estimate this next-state conditional by counting transitions between pairs of delay states.

Assume row $p_i$ has a Dirichlet prior $p_{i}\sim \mathrm{Dir}(\alpha)$,
and let $c_{ij}$ be the number of observed transitions $i\to j$ seen within the context window, 
\[
    N_i = \sum_j c_{ij}.
\]
Then the posterior mean is
\begin{equation}
    \widehat P_{ij} = \frac{c_{ij}+\alpha_j}{N_i+\sum_j \alpha_j}.
\label{eq:posterior}
\end{equation}
For large $N_i$, the Dirichlet prior contributes only minor corrections, so $\widehat p_i$ is asymptotically the empirical row-frequency estimator.

\vspace{1em}\noindent\textbf{Scaling of KL divergence.} Let the distribution $\v{p}_i \equiv \{p_{ij}\}_{j=1}^M$ describe the $i^{th}$ row of the true transition matrix. The row $\v{p}_i$ has an effective support size $M_i \leq M$, corresponding to the number of outgoing states reachable from state $i$. Let $\hat{\v{p}}_i \equiv \{\hat{p}_{ij}\}_{j=1}^M$ correspond to an estimator of the true $\v{p}_{i}$ derived from observing $N_i$ transitions in the context originating from state $s_i$. Assuming $\abs{\hat{p}_{ij}  -  p_{ij}} \ll 1$ for all $j$, a second-order Taylor expansion of the KL divergence gives
\begin{equation}
    D_\text{KL}(\v{p}_i\|\hat{\v{p}}_i)
    \approx
    \frac{1}{2}\sum_j \frac{(\hat{p}_{ij}-p_{ij})^2}{p_{ij}}.
\end{equation}
The expected value of the KL divergence over many samplings of set $\hat{\v{p}}_i$ is
\begin{equation}
    \mathbb{E}\!\left[D_\text{KL}(\v{p}_i\|\hat{\v{p}}_i)\right]
    \approx
    \frac{1}{2}\sum_j \frac{\mathbb{E}\!\left[(\hat{p}_{ij}-p_{ij})^2\right]}{p_{ij}}.
    \label{eq:expectedkl}
\end{equation}
We treat $\hat{\v{p}}_i$ as a random variable with multinomial statistics due to count-based estimation. The variance of a given element is given by
\begin{equation}
    \mathbb{E}\!\left[(\hat{p}_{ij}-p_{ij})^2\right]
    =
    \text{Var}(\hat{p}_{ij})
    =
    \dfrac{p_{ij}(1 - p_{ij})}{N_i}
\end{equation}
We substitute this into \eqref{eq:expectedkl}
\begin{equation}
    \mathbb{E}\!\left[D_\text{KL}(\v{p}_i\|\hat{\v{p}}_i)\right]
    \approx
    \frac{1}{2}\sum_j \frac{1-p_{ij}}{N_i}.
    \label{eq:expectedkl}
\end{equation}

\begin{equation}
    \mathbb{E}\!\left[D_\text{KL}(\v{p}_i\|\hat{\v{p}}_i)\right]
    \approx
    \frac{M_i-1}{2N_i}.
\label{eq:rowwise}
\end{equation}
Thus the expected error of a single row of the estimated transition operator scales as $\mathcal{O}(N_{i}^{-1})$, where $N_i$ is the number of transitions originating from state $i$ observed in the context.

We assume the process $p_{ij}$ is stationary with invariant distribution $\boldsymbol{\pi}^{(0)}$, and mixing is strong enough that over a context of length $C$ the number of expected counts of each state is given by the invariant distribution,
\[
    N_i \approx C\pi^{(0)}_i.
\]

Thus, once the delay embedding is large enough to make the process approximately Markov, increasing context length improves estimation of the transfer operator as $1/C$.

\vspace{1em}\noindent\textbf{Scaling laws for the spectrum of the transfer operator.} We describe approximated transfer operator $\hat{p} \in \mathbb{R}^{M \times M}$ as the ground truth transfer operator $p$ and an unknown error matrix $e \in \mathbb{R}^{M \times M}$, $\hat p = p + e$. We estimate rows of $\hat{p}$ based on $\mathcal{O}(C)$ effective samples overall, and so concentration gives
\begin{equation}
    \|E\| = \mathcal{O}(C^{-1/2}),
\end{equation}
up to dimension and occupancy factors.

If $\lambda$ is an isolated eigenvalue of $P$, then Weyl's bounds from eigenvalue perturbation theory correspond to
\begin{equation}
    |\hat\lambda-\lambda| = O(C^{-1/2}).
\end{equation}
Likewise, if $\boldsymbol{\pi}^{(k)} \in \mathbb{R}^D$ denotes an eigenvector associated with the eigenvalue $\lambda_i$ separated by spectral gap $\mathrm{gap}_i = \min_{j\neq i}\abs{\lambda_i - \lambda_j}$, then Davis-Kahan bounds imply
\begin{equation}
\|\sin\theta(\boldsymbol{\pi}^{(k)}, \hat{\boldsymbol{\pi}}^{(k)})\|
=
\mathcal{O}\!\left(\frac{C^{-1/2}}{\mathrm{gap}_i}\right).
\end{equation}
where $\theta(\boldsymbol{\pi}^{(k)}, \hat{\boldsymbol{\pi}}^{(k)}) \in \mathbb{R}$ denotes the angle between the true ($\boldsymbol{\pi}^{(i)}$) and estimated ($\boldsymbol{\hat{\pi}}^{(i)}$) modes of the transfer operator. The invariant distribution is the left eigenvector at eigenvalue $1$. Perturbation theory thus implies
\begin{equation}
\|\hat{\boldsymbol{\pi}}-\boldsymbol{\pi}\|
=
    \mathcal{O}\!\left(\frac{C^{-1/2}}{1-|\lambda_2|}\right),
\end{equation}
up to conditioning constants, where $1-|\lambda_2|$ is the mixing gap. In the small-error regime, the KL divergence is locally quadratic, so
\begin{equation}
    D_{\mathrm{KL}}(\boldsymbol{\pi}\|\hat{\boldsymbol{\pi}}) = \mathcal{O}(C^{-1}).
\end{equation}
Hence improved invariant-measure estimation is a direct consequence of improved finite-context operator estimation.

\putbib[cites]
\end{bibunit}
\end{document}